\documentclass[runningheads]{llncs}
\usepackage{graphicx}
\usepackage{subfigure}
\usepackage{amsmath}
\usepackage{mathabx}
\usepackage{booktabs}
\usepackage{algorithm}
\usepackage{algorithmicx}
\usepackage{algpseudocode}
\usepackage{xcolor}
\usepackage{cleveref}
\usepackage{url}
\usepackage{soul}
\newcommand{\x}{{\bf x}}
\newcommand{\w}{{\bf w}}
\newcommand{\SaDe}{{\em SaDe}}

\begin{document}
\title{SaDe: Learning Models that Provably Satisfy Domain Constraints}
%
%
\author{Kshitij Goyal\inst{1} \and
Sebastijan Dumancic\inst{2} \and
Hendrik Blockeel\inst{1}}
%
\authorrunning{K. Goyal et al.}
%
\institute{KU Leuven \and TU Delft}

%
\maketitle              
\begin{abstract}
In many real world applications of machine learning, models have to meet certain domain-based requirements that can be expressed as constraints (e.g., safety-critical constraints in autonomous driving systems).
Such constraints are often handled by including them in a regularization term, while learning a model. This approach, however, does not guarantee 100\% satisfaction of the constraints: it only reduces violations of the constraints on the training set rather than ensuring that the predictions by the model will {\em always} adhere to them.
In this paper, we present a framework for learning models that {\em provably} fulfill the constraints under {\em all} circumstances (i.e., also on unseen data).
To achieve this, we cast learning as a maximum satisfiability problem, and solve it using a novel {\em SaDe} algorithm that combines constraint satisfaction with gradient descent.
We compare our method against regularization based baselines on linear models and show that our method is capable of enforcing different types of domain constraints effectively on unseen data, without sacrificing predictive performance.
\end{abstract}

\section{Introduction}
There is increasing interest in using machine-learned models in contexts where strict requirements exist about the model's behavior. For instance, in a criminal sentencing context, a fairness constraint might express that all else being equal, two people of a different ethnicity should have an equal probability of ending up in jail \cite{barocas2017fairness}.  In another example, when automating parts of an aircraft system, the model may be required to satisfy certain safety-critical requirements \cite{katz2019marabou}. We call such requirements {\em domain constraints}, as they constrain the behavior of the learned model over its whole domain. 

Machine learning methods often deal with such constraints by including them in the cost function they optimize (e.g., in a regularization term) \cite{berner2021modern}. This approach has the effect of {\em encouraging} the learner to learn a model that satisfies the imposed constraints on the {\em training} data, but it does not
{\em guarantee} that the learned model satisfies the constraints over the whole {\em input space}.
While this may be good enough when the constraints are intended to help the learner obtain better models from less data \cite{diligenti2017semantic,xu2018semantic}, it is insufficient for applications where constraint satisfaction is imperative under all circumstances (such as safety-critical systems).

For this reason, research had been conducted on approaches that can guarantee constraint satisfaction even on unseen data, as in \cite{sivaraman2020counterexample} where a counter-example guided approach is used to enforce monotonicity constraints, and \cite{hoernle2021multiplexnet} where a multiplexer layer is used as the output layer in a neural network to enforce domain constraints.
The existing literature, however, still lacks a general approach that can be used to enforce a variety of domain constraints on different learning problems with provable guarantees.

In this work, we present a framework for learning parametric models that are guaranteed to satisfy domain constraints.
Rather than including the constraints in a cost function and using a standard learning approach, 
the machine learning problem is cast into a constraint satisfaction problem \cite{rossi:cphandbook}, more specifically a Maximum Satisfiability Modulo Theories (MaxSMT) problem \cite{fu2006solving}.
Domain constraints are formulated as hard constraints, which must provably be satisfied, whereas the model’s fit with the training data is evaluated using soft constraints, of which as many as possible should be satisfied. Thus, a model is found that optimally fits the data within the hard constraints imposed by the user.


Unfortunately, solving the obtained MaxSMT problem does not scale beyond a few dozen training instances.
To resolve this, we propose {\em Satisfiability Descent} (\SaDe), a variant of gradient descent \cite{gori2017machine} in which each step consists of solving a small MaxSMT problem to find a local optimum in the general direction of the negative gradient (rather than moving in the exact direction) that satisfies all domain constraints.
We show experimentally that \SaDe\ scales to realistically-sized datasets and that it finds models with similar performance as other learners while guaranteeing satisfaction of all domain constraints.

In Sections 2--4, we consecutively introduce preliminaries, the MaxSMT-based approach, and \SaDe. We position SaDe with respect to related work in Section 5 and present the empirical evaluation in Section 6.  Section 7 concludes.

\section{Preliminaries}

We systematically use boldface for vectors and italics for their components, e.g., ${\bf x} = (x_1, x_2, \ldots, x_n)$.

\subsection{SAT, MaxSAT, SMT, MaxSMT, COP}
\label{sec:prelim}

Let {\bf w} denote a vector of decision variables, and $C_i$ a (hard or soft) constraint, i.e., a boolean function of {\bf w}.  We say that \w\ satisfies $C_i$ if and only if $C_i(\w)$ returns true. We call {\bf w} {\em admissible} if it satisfies all hard constraints. We can then distinguish the following types of problems:
\begin{itemize}
\item SAT : given a set of constraints $C_i(\w)$, $i=1, \ldots, k$, determine whether an admissible $\w$ exists
\item MaxSAT: given a set of hard constraints $\mathcal{H}$ and a set of soft constraints $\mathcal{S}$, find an admissible $\w$ that satisfies as many $C_i \in \mathcal{S}$ as possible  
\item SMT: Satisfiability modulo theories: this setting is identical to SAT, except that not only logical reasoning is used, but also a theory on the domain of $\w$. For instance, an SMT solver knows $x<y \land y<x$ is unsatisfiable; a SAT solver does not, because it does not know the meaning of $<$.
\item MaxSMT: similar to MaxSAT, but satisfiability is determined modulo theories
\item COP: constraint optimization: given a set of constraints and a function $f$, find the admissible $\w$ with smallest $f(\w)$ (among all admissible $\w$).
\end{itemize}

MaxSAT reduces to SAT in the sense that any MaxSAT problem can be solved by iteratively solving SAT problems.
The Fu-Malik algorithm \cite{fu2006solving} is an example of such an approach. Similarly, MaxSMT reduces to SMT. This implies that if we know how to solve the SMT problem for a particular type of theory, we can automatically solve the corresponding MaxSMT problem.

COP problems can be approximately solved by turning them into a MaxSMT problem, as follows: make all the original constraints hard constraints, and add soft constraints of the form $f(\w)<c_i$, $i=1, \ldots, k$ where $f$ is the function to be minimized and $c_i < c_{i+1}$. The solution is approximate in the sense that if $\hat{\w}$ is the returned solution and $\w^*$ the actual optimum, $f(\hat{\w})-f(\w^*) \leq c_i - c_{i-1}$ for some $i$ (i.e., closer thresholds guarantee a better solution).

Due to these properties, the solving power of SMT solvers can be lifted towards (approximate) constrained optimization. This is a key insight behind our approach. 

\subsection{Universally quantified constraints}

Constraint solvers assume a finite set of constraints. Different solvers may use different languages in which these constraints can be expressed. Some solvers allow for the constraints to contain universal quantifiers, for instance (expressing monotonicity of $f$ in some input variable $x_i$):
\begin{equation}
    \forall \x, \x' \in \mathcal{X}: x_i \leq x'_i \land (\forall j \neq i: x_j = x'_j) \implies f_{\w}(\x) \leq f_{\w}(\x')
\label{eq:monot}
\end{equation} 

When the universal quantification is over a variable with finite domain, such a constraint can always be handled by {\em grounding} it: making a separate copy for each value of the domain. For infinite domains, however, this is not possible. SMT solvers typically handle such cases by turning the quantified variable into a decision variable, and then, through reasoning, eliminating the quantifier.
E.g., the constraint $\forall x>0: f(x)>0$ with $f(x)=ax+b$ cannot be turned into a finite set of constraints of the form $f(1)>0, f(2)>0, \ldots$ but an SMT system can deduce an equivalent constraint on the model parameters, namely $a>0 \land b>0$. 





With this approach, the extent to which universally quantified constraints can be handled clearly depends on the strength of the mathematical reasoning engine.  In this work we use Z3 \cite{de2008z3}, one of the more powerful systems in this respect. Z3 implements an SMT(NRA) solver: an SMT solver that can reason with non-linear equations (NRA = Nonlinear Real Arithmetic), and uses this to deal with universally quantified constraints. The ability to handle nonlinear functions is crucial for our approach, even when learning linear models. That is because turning the quantified variable $x$ into a decision variable gives rise to formulas in which products of decision variables occur.  E.g., we typically think of $ax+b$ as linear because we think of $a$ and $b$ as constants, but to the solver, $a$, $x$, and $b$ are all variables, and $ax+b$ is no more linear than $f(x,y,z)=xy+z$.

\section{From constrained parametric machine learning to MaxSMT}

In this section, we propose a framework for formulating supervised parametric machine learning as a MaxSMT problem. 

\subsection{The learning problem}

We consider the following learning problem:
 
\begin{definition}{\textbf{Learning problem.}}
Given a training set $D \subseteq \mathcal{X} \bigtimes \mathcal{Y}$, a set of constraints $\mathcal{K}$, a loss function $\mathcal{L}$, and a hypothesis space containing functions $f_\w: \mathcal{X} \rightarrow \mathcal{Y}$; find $\w$ such that $f_{\w}$ provably satisfies constraints $\mathcal{K}$ and $\mathcal{L}(f_\w, D)$ is minimal among all such $f_{\w}$.
\label{def:2}
\end{definition}

The language in which the constraints in $\mathcal{K}$ are expressed is essentially a subset of first-order logic. Formulas can contain universal quantification ($\forall$) over known sets, notably the training set $D$ and the input space $\mathcal{X}$; arithmetic operators are defined, as well as operators that extract a component from a tuple; and the formula can refer to the function $f_{\w}$ for a given value of \w. The variable $\w$ is free: depending on its value, the function $f_{\w}$ either fulfills or violates the constraint $\mathcal{K}$.
Examples of expressible constraints are monotonicity (see Equation \ref{eq:monot}) and conditional bounds, e.g. (inspired by safety-critical applications \cite{katz2017reluplex}): $\forall \x \in \mathcal{X}: x_i > a \implies f_{\w}(\x) > 0$. 
For binary classification problems, we assume a real value prediction with $f_\w$, which is then translated into a binary decision using a sigmoid function.

\subsection{Translation to MaxSMT}

Though the above-defined learning problem looks quite standard, we could not find any constraint-based optimization approach that can handle it, among many we considered (which includes constraint programming and mixed integer linear programming). This finding is actually consistent with earlier work \cite{dumancic2020automated}. Solving the problem required a combination of the ability to handle universal quantification over continuous domains, non-linear real arithmetic, and optimization that no system offers. The easiest way out was to drop the ``optimization'' aspect and reduce the COP problem to a MaxSMT(NRA) and ultimately an SMT(NRA) approach. We have implemented such an approach on top of the Z3 solver.

Z3 contains algorithms for solving COP problems directly, but these cannot deal with universally quantified constraints. 
We therefore convert the COP problem to MaxSMT(NRA) in a way that is similar to the procedure explained in section \ref{sec:prelim}. 
We approximately encode the loss function $\mathcal{L}$ using soft constraints that we call {\em decision constraints}.
Decision constraints impose a certain quality of fit on $f_{\w}$. 
They are typically of the form $C(f_\w(\x), y)$, with $C$ some condition that is fulfilled when $f_\w(\x)$ is ``sufficiently consistent’’ with the observed $y$, for a given $({\bf x}, y) \in D$. In this paper, we consider two different forms of $C$, depending on whether $y$ is boolean (binary classification) or numerical (regression).

For regression, decision constraints are of the following form:
\begin{equation*}
    y - e \leq f_\w(\x) \leq y + e
\end{equation*}
with $e$ some threshold. Multiple such constraints, each with a different threshold, can be introduced for each data point: the closer $f_\w(\x)$ is to $y$, the more such constraints are satisfied for the data point $(\x,y)$.  Depending on the context, the threshold $e$ can be set relative to the value of $y$, e.g., $e = 0.1*y_{max}$ where $y_{max} = \max_D |y|$.

For binary classification, we assume that the sign of $f_\w(\x)$ indicates the class, and its magnitude indicates the model's certainty about the prediction.  Hence, we use decision constraints of the following form:
\begin{eqnarray*}
& &  f_\w(\x) > \tau   \qquad \text{if $y = 1$}\\
& &  f_\w(\x) < -\tau   \qquad \text{if $y = -1$}
\end{eqnarray*}
for some threshold $\tau$. Again, multiple such constraints can be used, with varying thresholds.

\subsection{Solving the MaxSMT problem}

Z3 natively support a number of MaxSMT(NRA) algorithms, but this module of Z3 does not support universal quantifiers over real variables. It does support such quantifiers for SMT(NRA). We therefore made our own MaxSMT(NRA) solver by implementing the Fu-Malik algorithm \cite{fu2006solving} on top of the SMT(NRA) solver that is provided in Z3. The Fu-Malik algorithm solves MaxSAT problems iteratively: it consecutively identifies minimal sets of constraints that are jointly unsatisfiable and relaxes the problem by allowing exactly one of these to be violated; it keeps doing this until the relaxed problem is satisfiable. 



One more change was needed to make this approach work. With our experiments, we realise that unbounded continuous domains make the learning very slow with Z3. 
To mitigate this issue, our approach assumes a bounded input space, where vectors $\bf{l}$ and $\bf{u}$ exist such that $l_i < x_i < u_i$ for all $i$, for all $\x \in \mathcal{X}$.
These bounds can be provided by the user, or we can use as defaults $l_i=\min_D (x_i)$ and $u_i = \max_D(x_i)$. 
Enforcing such bounds is also practical: a continuous feature in a machine learning task always has a range of values it can realistically take.
For example, the {\it age} of a person can only be in the range of [0, 150] and a value of, say 10000, is unrealistic.
Hence, it's sensible to enforce the domain constraints in such realistic ranges, and the most straight-forward way to get these ranges is the training data itself.
In the remainder of this paper, when we have quantification over $\mathcal{X}$, it should be kept in mind that we actually assume a bounded $\mathcal{X}$.

\begin{figure}[t]
  \centering
  \includegraphics[scale=0.11]{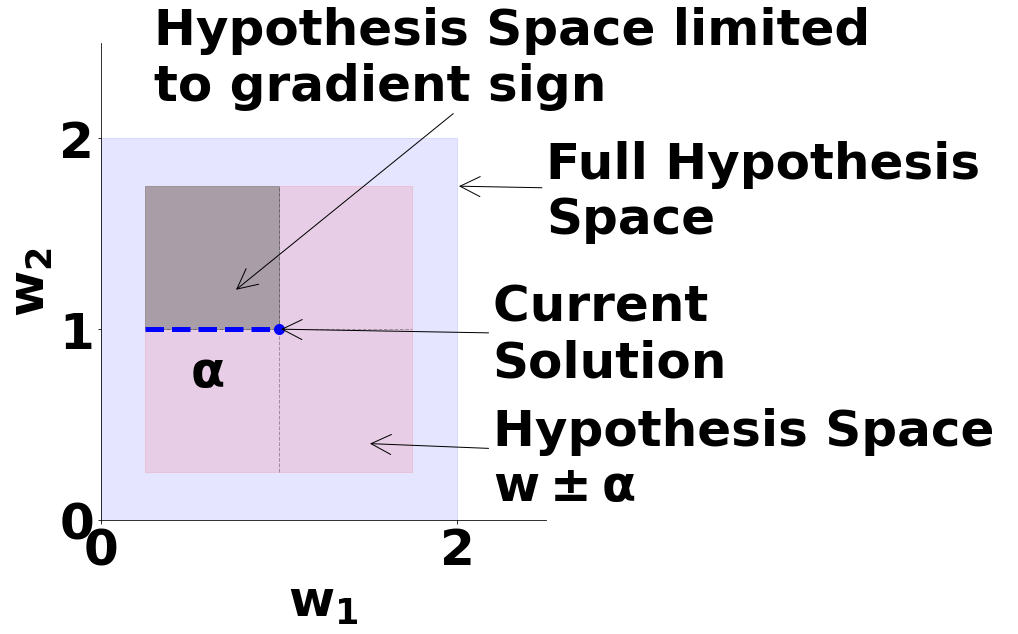}
  \includegraphics[scale=0.16]{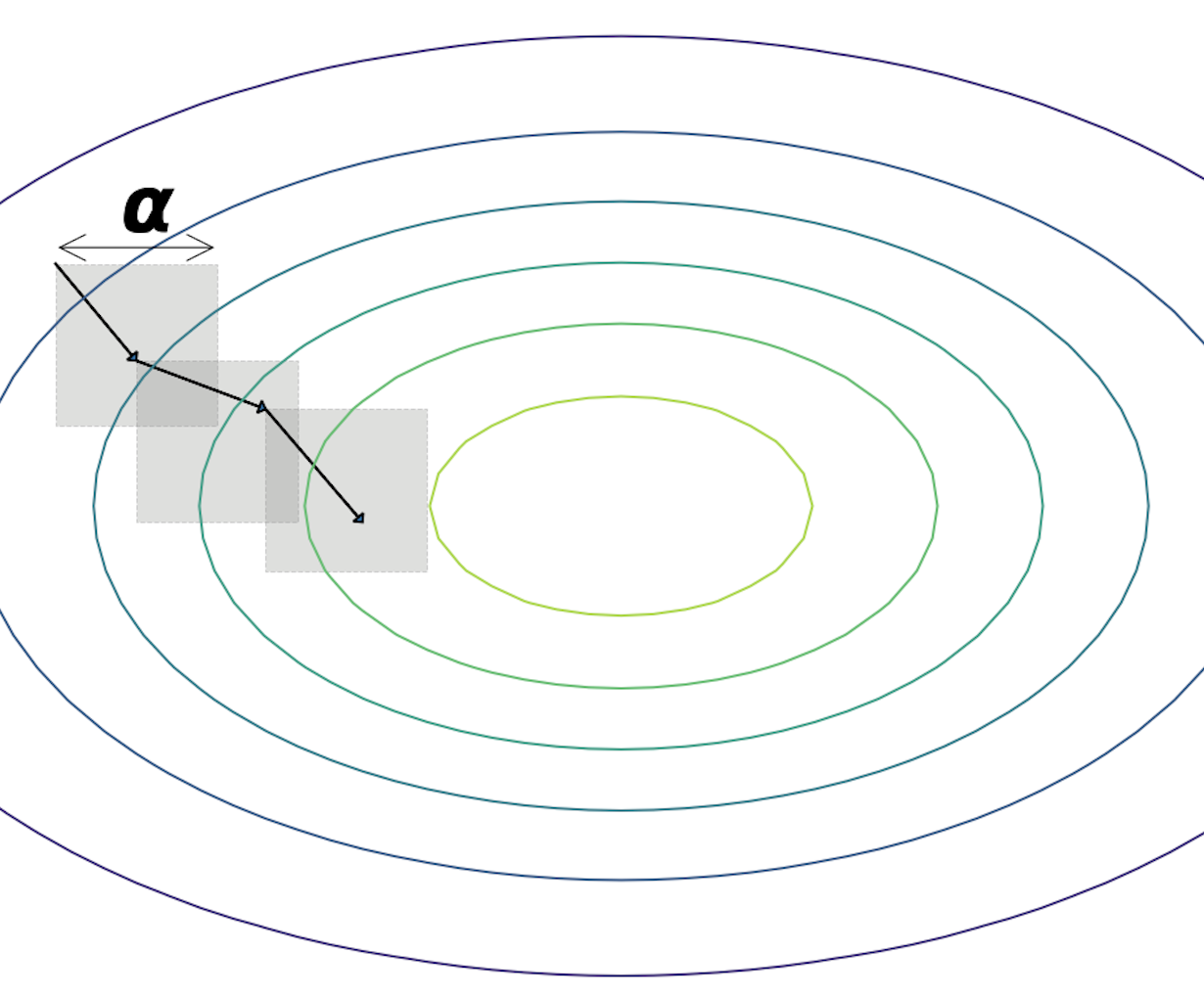}\\
  \caption{An Illustration of the SaDe algorithm for two parameters. 
  \textbf{Left figure:} For the current solution $\{w_1 = 1, w_2 = 1\}$, assuming the gradients $\{\frac{\delta \mathcal{L}}{\delta w_1} >0, \frac{\delta \mathcal{L}}{\delta w_2} <0\}$, the hypothesis space for the next solution is the grey quadrant;
  \textbf{Right Figure:} Every grey quadrant represents the search space from one iteration to the next (maximal step size: $\alpha$), decided by the gradients of the loss. 
  }
  \label{fig:sade}
\end{figure}

\section{SaDe: Satisfiability Descent}

The approach explained above is straightforward and intuitive, but unfortunately not scalable.
In our preliminary experiments, we observed that the above approach became prohibitively slow beyond a few dozen training instances.
The reason for this is the combinatorial nature of MaxSMT: increasing the number of instances, and consequently the number of soft constraints, makes the problem exponentially more complex. 

To overcome this limitation, we have devised an algorithm called {\em Satisfiability Descent} (\SaDe).  
Essentially, \textit{SaDe} just performs gradient descent, like other learning algorithms.  However, it cannot simply ``take a step in the direction of the negative gradient'', as the point where it arrives may not be admissible.  Instead, the MaxSMT procedure is used to find an admissible point near the point that gradient descent would lead to. 
More precisely: Let $\mathcal{L}$ be the loss measured on the whole training set, and $\mathcal{L}_B$ the number of violated constraints in a {\em batch}, a small subset of the training data (small enough that MaxSMT is feasible). The gradient descent principle makes \SaDe\ move in the direction of a local optimum of $\mathcal{L}$, while the MaxSMT procedure makes sure the next point is admissible and minimizes $\mathcal{L}_B$ in a local region. A motivating assumption behind minimizing $\mathcal{L}_B$ is that the loss function correlates with the number of violations. This is true for commonly used losses (e.g., mean squared error, cross-entropy loss) and for the decision constraints introduced here.


\begin{algorithm}[t]
\caption{SaDe}\label{sade}
\hspace*{\algorithmicindent} \textbf{input:} training data $D$, domain constraints $\mathbf{\mathcal{K}}$, batch size $b$, number of epochs $e$, \hspace*{\algorithmicindent} loss $\mathcal{L}$, maximal step size $\boldsymbol\alpha$ \\
\hspace*{\algorithmicindent} \textbf{output:} optimal parameter values $\w$ 
\begin{algorithmic}[1]
\State $W = \{ \}$, $\mathcal{H} = \mathcal{K}$, ${\bf g} = \mbox{undefined}$
\State partition $D$ into batches of size $b$
\While{stop\_criterion not fulfilled}
    \For{each batch $B$ in $D$}
        \State $\mathcal{S}$ = DECISION\_CONSTRAINTS($B$)
        \State $sol = MaxSMT(\mathcal{S}, \mathcal{H})$
        \If{$sol.label \text{ is } SAT$}
            \State $\hat{\w} = sol.params$
            \State $W = W \cup \{ \hat{\w} \}$
            \State ${\bf g} = \nabla \mathcal{L}(\hat{\w})$
        \ElsIf{{\bf g} is defined}
            \State {\bf g} = - {\bf g}
        \EndIf
        \If{{\bf g} is defined}
            \State $\mathcal{H}$ = $\mathbf{\mathcal{K}} \cup \{\w \in Box(\hat{\w}, \hat{\w}- \alpha \cdot \mbox{sgn}({\bf g})) \}$
        \EndIf
    \EndFor
\EndWhile
\State \Return $\underset{\hat{\w} \in W}{\arg\min} (\mathcal{L}(D, \hat{\w}))$
\end{algorithmic}
\end{algorithm}

Algorithm \ref{sade} shows pseudocode for \SaDe. The algorithm runs for multiple epochs, each time processing all batches sequentially. It starts with finding a solution for the first batch; this solution must satisfy all domain constraints (stored in the set of hard constraints $\mathcal{H}$) and as many soft constraints ($\mathcal{S})$ as possible. The solution is stored in an object {\em sol} with a field {\em label} that equals SAT if the problem is satisfiable and a field {\em params} that in that case contains the solution. It adds this solution to a set $W$, computes the gradient of the loss function at this point, stores this gradient in variable {\bf g}, and extends the hard constraints with a ``box'' constraint, which states that the next solution must be inside the axis-parallel box defined by $\hat{\w}$ and $\hat{\w} - \alpha \cdot \mbox{sgn}({\bf g})$, where the sign function is applied component-wise to a vector.\footnote{We use a modified sign function where sgn$(0)=1$, so that the box never reduces to a lower-dimensional box.} In other words, each $w_i$ will be confined to the interval $[\hat{w}_i, \hat{w}_i+\alpha]$ or $[\hat{w}_i-\alpha, \hat{w}_i]$, depending on the sign of $g_i$. This forces the algorithm to move, not in the exact direction of the negative gradient, but in a direction that lies in the same orthant; see Fig.~\ref{fig:sade} for an illustration.

The box constraint may render the problem unsatisfiable. In that case, the algorithm takes a step back (lines 12, 15) and continues with the next batch from there.
We call this a \textit{restart}. Such a restart is also made when the solver does not find a solution within reasonable time (5 seconds, in our implementation).

Each intermediate solution is stored (line 9) and the one that minimizes the loss is returned as the final solution (line 19).
\textit{SaDe} runs until some stopping criterion is fulfilled. Our current implementation checks every $100^{th}$ iteration (starting from the 400th) whether the loss is improved by at least 2\%, compared to 200 iterations ago. 
If the improvement is less than $2\%$, or if a maximum number of iterations is reached, the process stops.
This criterion is not crucial to the algorithm and can be replaced with another one.

\section{Related Work}

There is a substantial body of work on imposing constraints on machine-learned models.  We  distinguish {\em syntactic} constraints, which constrain the structure of the model (e.g., maximal depth, for a decision tree), and {\em semantic} constraints, which constrain its behavior. The first type is easier to impose, and can serve as a proxy for the second. An example are the feature interaction constraints  in XGBoost\footnote{\url{https://xgboost.readthedocs.io/en/stable/tutorials/feature_interaction_constraint.html}}: avoiding co-occurrence of two attributes in the same tree precludes interaction (in the statistical sense) between them. In neural networks, the architecture of the network can be chosen so that it enforces certain semantic constraints \cite{berner2021modern,hoernle2021multiplexnet}.
What semantic constraints can be imposed through syntactic constraints depends on the model format, but in general, the set is limited and ad hoc.  E.g., the multiplexnet \cite{hoernle2021multiplexnet} is relatively versatile, but still limited to quantifier-free formulas.

Multiple approaches have been proposed that enforce constraints through regularization (e.g., \cite{diligenti2017semantic,fischer2019dl2,xu2018semantic}).
These approaches typically allow for a much wider range of constraints to be  expressed. However, they treat these constraints as soft constraints and cannot handle universal quantification over the domain.

Convex optimization based methods (e.g., support vector machines) inherently include hard constraints in the optimization task. Given that they already deal with such constraints, one can just as well add more constraints to express domain knowledge. However, the type of  constraints that can be expressed is again limited; e.g., no quantifiers over continuous domains can be used.

Methods that rely on combinatorial optimization are closest to our work. Such methods have been proposed for decision trees (e.g., \cite{demirovic2020murtree,hu,verhaeghe2019learning,verwer2019learning}), but typically with syntactic constraints (e.g., find an optimal decision tree of depth at most 5). 
There are some optimal decision tree methods that impose semantic constraints~\cite{aghaei:optimalandfairdt,vos2021robust}, but without guaranteed constraint satisfaction on unseen data.
\cite{sivaraman2020counterexample} proposes a counter-example guided approach to enforce monotonicity constraints for all possible unseen instances, but lacks a general framework for other types of constraints.
\cite{manhaeve2018deepproblog} proposes an approach to include logical constraints in neural network training using ProbLog, but is limited to classification problems and doesn't guarantee constraint satisfaction.
MaxSAT has previously been used in various machine learning tasks, like Bayesian networks~\cite{berg2019applications,cussens2012bayesian}, interpretable classification rules \cite{malioutov2018mlic} and optimal decision sets \cite{yu2021learning}.
These approaches, however, learn in a discrete domain and do not support imposing domain constraints.
To the best of our knowledge, ours is the first work that learns parametric models (with a continuous domain) in a MaxSMT framework.

Apart from all these approaches, our work also relates to work on verification of learned models, such as neural nets \cite{huang2017safety,katz2019marabou,singh2018boosting} and tree based models \cite{einziger2019verifying,chen2019robustness,devos2021versatile}. That work uses similar methods, but merely checks that a learned model meets certain requirements, rather than enforcing this through the learner.

\section{Experimental evaluation}

The fact that SaDe {\em guarantees} compliance with domain constraints crucially distinguishes it from other systems, such as regularization-based methods.  Even then, a number of questions can be raised:
\begin{itemize}
    \item[{\bf Q1}] Does it matter in practice?  Perhaps other methods often learn admissible models anyway, even if they do not guarantee it.
    \item[{\bf Q2}] Does this affect predictive performance?
    \item[{\bf Q3}] What is the cost in terms of learning efficiency?
\end{itemize}
We address these questions empirically. We first describe the use-cases, then the evaluation methodology, and finally the results.

\subsection{Use-cases}

SaDe supports any constraint and model expressible in the SMT-LIB language \cite{ranise2006smt}; consequently, SaDe supports any machine learning task and setting that can be expressed in the same language.
To demonstrate this ability, we design three use-cases with different tasks and settings: a binary classification problem of loan prediction, a multi-class classification problem of music genre prediction, and a multi-target regression problem of expense prediction.
Despite this variety, the use-cases have the following in common: they include universally quantified domain constraints, and some of the input data violate these constraints. The later is motivated by the fact that learning robust models is more challenging, but also more useful, when training data may violate constraints (e.g., data may contain undesirable bias that we explicitly do not want to model).

For readability, we use names rather than numerical indices for tuple  components; e.g., {\em artist}({\bf x}) refers to the component of {\bf x} that indicates the artist. 

Our first use-case is a {\bf music genre identification} problem.
Data, consisting of 793 songs, comes from a music streaming company 
Tunify\footnote{\url{https://www.tunify.com/en-gb/}}.
Each song is represented by 13 features and belongs to one of 5 classes: \textit{rock, pop, classical, electronic, metal}. 
This is a {\it multi-class classification problem}, which we convert to several binary classification problems using a one-versus-all approach.
The final prediction for a new instance is the class corresponding to the binary classifier with the highest confidence. We impose the domain constraint requiring that \textit{a Beatles song can only be classified to either Pop or Rock}, encoded as:
\begin{multline*}
     \forall \x \in \mathcal{X}: artist(\x) = \textit{The Beatles} \implies ((rock(f_{\w}(\x)) > 0 \lor pop(f_{\w}(\x)) > 0)  \land \\ classical(f_{\w}(\x)) < 0 \land electronic(f_{\w}(\x)) < 0 \land metal(f_{\w}(\x)) < 0)
\end{multline*}
The dataset contains $60$ violations.
Our second use case is the {\bf loan approval problem}\footnote{\url{https://www.kaggle.com/altruistdelhite04/loan-prediction-problem-dataset}}.
The data consists of 614 instances with 6 categorical and 5 numerical features.
This is a binary classification problem: predict whether the loan should be approved or not.
We impose the domain constraint requiring that {\it everyone with no credit history (ch) and income lower than 5000\$ should be denied a loan}, which is encoded as:
\begin{equation*}
     \forall \x \in \mathcal{X}: ch(\x) = 0 \land income(\x) < 5000 \implies f_{\w}(\x) < 0
\end{equation*}
\noindent The dataset contains $30$ violations.
Our final use-case is the {\bf expense prediction problem}\footnote{\url{https://www.kaggle.com/grosvenpaul/family-income-and-expenditure}}. which consists of predicting multiple types of expenses for a household. 
The data consists of 1000 instances with 5 target expenses and 13 predictors.
This is a \textit{multi-target regression problem} which is converted into a collection of single target regression problems, one for each target (we use $exp$ to represent a target in the expressions below).
We enforce two domain constraints requiring that {\it the sum of all expenses must be smaller than the household income and going-out expense must not be more than 5\% of the household income.}
Domain constraints are encoded as
\begin{multline*}
 \forall \x \in \mathcal{X}: (\sum_{exp} exp(f_{\w}(\x)) \leq income(\x)) \bigwedge  (going\_out(f_{\w}(\x)) \leq 0.05*income(\x))
\end{multline*}
\noindent The dataset contains $862$ violations.
\subsection{Evaluation methodology}

\subsubsection{Evaluation metrics:}
To answer question {\bf Q1}, we need to measure ``reliability'': how certain are we that the model will not violate any constraints? To define a measure for this, we consider {\em counterexamples}: instances for which the model's prediction violates at least one domain constraint. We define the {\bf adversity index (AdI)} as the percentage of training instances for which a counterexample can be constructed in an $l_{\infty}$ ball with radius $\delta$ centered around the instance. Note that the counterexample need not be part of the training set itself, but it must be similar to a training instance. This avoids the construction of ``unrealistic'' counterexamples that are totally different from anything ever seen and might not exist in practice. Counterexamples are constructed by simply using the SMT solver.

Measuring predictive performance (for question {\bf Q2}) requires some care. We use accuracy (for classification) and mean squared error (MSE) (for regression) as performance metrics. But we should not simply compute these on the whole test set: some labels in the test set may violate the constraints and in such cases the model should explicitly {\em not} predict the same value. It is not known, however, what value should be predicted instead. For this reason, predictive performance is computed on the subset of the test data that satisfies the constraints.

\subsubsection{Evaluation procedure:}
We use nested 5-fold cross validation, in which the inner cross-validation is used to select the hyper-parameters.
SaDe's hyper-parameters are the maximal step size $\alpha$, which is selected from $\{0.5, 1, 2\}$ and the thresholds used to define the decision constraints.
For classification, these thresholds are selected from $\{[0, 1], [0, 1, 2],[1, 2]\}$; for regression, they are $c*\max_D |y|$ with $c$ selected from $\{[0.1], [0.1, 0.2], [0.1, 0.2, 0.3]\}$.
The model class that SaDe uses for $f_{\bf w}$ is linear models, and the loss function is cross-entropy for classification, and sum of mean squared error (MSE) over all target variables for regression.
The regularisation-based baselines that we compare SaDe to use the same model class and loss functions. They have one hyper-parameter, $\lambda$, which is the standard trade-off between the loss function and regularisation term ($loss + \lambda \cdot regularisation$).
The value of $\lambda$ that leads to minimum number of violations on a validation set is selected via cross-validation. 
All the features are scaled to $[0, 1]$.
The experiments are repeated 10 times, with each model being trained for 10 epochs and a batch size of 5.
We use the SMT(NRA) solver z3 (version 4.8.10) for SaDe and an Intel(R) Xeon(R) Gold 6230R CPU @ 2.10GHz machine with 256 GB RAM.

\subsubsection{Baselines:}

For classification, we compare SaDe to Semantic-Based Regularisation (SBR) \cite{diligenti2017semantic} and Semantic Loss (SL) \cite{xu2018semantic} regularisation-based approaches.
Note that these do not support universally quantified constraint over infinite domains: they simply ground such constraints over the training examples.  
For regression, we compare SaDe with a baseline model where we regularize the mean squared error loss with an additional penalty whenever the constraint is violated on the training data. E.g., for use case 3, this regularized loss is:
\begin{multline*}
    	\mathcal{L}_{R}= MSE + \lambda * \frac{1}{\|D\|} * \sum_{\x \in D}(\max(0, \sum_{exp}exp(f_{\w}(\x)) - income(\x)) \\
    	+ \max(0, going\_out(f_{\w}(\x)) - 0.05*income(\x)))
\end{multline*}
We will refer to this baseline as SBR in the remaining text.

For the classification task, we additionally consider a ``post-processing'' (PP) baseline: train the model without regard for any constraints; at prediction time, check whether the prediction violates a constraint, and if it does, change it. For classification, we assume that PP flips the prediction to the highest-scoring class that satisfies the domain constraint.  Note that, while PP provides a trivial way to enforce domain constraints at prediction time, it is not a generally applicable method: it requires that we know how to ``fix'' the prediction, which is not always the case (as will be illustrated for the regression use case in the next section).

\subsection{Results}

We consecutively interpret the experimental results in the light of the three research questions listed before.  The Post-Processing approach is discussed separately after that.

\setlength{\tabcolsep}{6pt}
\begin{table}[t]
    \centering
    \begin{tabular}{@{}lrrrrr@{}}
        \toprule
        \textbf{Use-case} & \textbf{Radius($\delta$)} &  \textbf{SaDe} &  \textbf{SBR} &  \textbf{SL} &  \textbf{PP} \\
        \midrule 
        Music Genre     & 0.01  & $0 \pm 0$ & $0.007 \pm 0.004$ & $0.007 \pm 0.005$ & $0 \pm 0$\\
                        & 0.1   & $0 \pm 0$ & $0.089 \pm 0.031$ & $0.024 \pm 0.014$ & $0 \pm 0$ \\
        \midrule 
        Loan Approval   & 0.01  & $0 \pm 0$ & $0 \pm 0$ & $0 \pm 0$ & $0 \pm 0$\\
                        & 0.1   & $0 \pm 0$ & $0 \pm 0$ & $0 \pm 0$ & $0 \pm 0$\\
        \midrule
        Expense Prediction  & 0.01  & $0 \pm 0$ & $0.056 \pm 0.009$ & - & -\\
                            & 0.1   & $0 \pm 0$ & $0.775 \pm 0.029$ & - & -\\
        \bottomrule
    \end{tabular}
    \caption{Adversity indices for all models. While it is not possible to construct a counter-example for SaDe models, regularisation-baselines are susceptible to them. As evident in the loan approval use-case, regularisation-based approach can occasionally result in reliable models that obey constraints, but that is not a rule }
    \label{tab:adi}
\end{table}

\subsubsection{Q1} Do other methods return inadmissible models? Table \ref{tab:adi} shows adversity indices for all models. The used values for $\delta$ are chosen to be small compared to the average $\ell_\infty$ distance between a pair of training instances ($0.89$ and $0.77$ in the Music and Expense datasets, respectively), so that the constructed counter-examples can be said to be similar to some training instances. The results indicate that regularisation-based approaches are highly sensitive to counter-examples (while SaDe, by construction, is not). For SBR models and a radius of $\delta = 0.01$, it is possible to construct a counter-example in the neighbourhood of 0.7\% and 5\% of instances in the Music Genre and Expense Prediction use-cases, respectively.
When the radius is increased to $\delta = 0.1$, it is possible to construct counter-examples in the neighbourhood of 9\% and 77\% of training instances in the Music Genre and Expense Prediction use-cases, respectively. SL seems slightly more robust, but counter-examples can still be found.


The loan approval use-case, on the other hand, shows that regularisation-based techniques can produce models that satisfy all constraints (no counter-examples could be constructed); they just do not guarantee it. It is not known under which conditions SBR and SL result in admissible models.

Overall, these results answer {\bf Q1 positively}: 
learners that do not {\em guarantee} that the learned models are admissible often return models that indeed are not.


\begin{table}[t]
    \centering
    \begin{tabular}{@{}lrrrr@{}}
        \toprule
        \textbf{Use-case} &  \textbf{SaDe} &  \textbf{SBR} &  \textbf{SL} &  \textbf{PP} \\
        \midrule 
        Music (accuracy)    & $80.76 \pm 5.15$ &  $82.94 \pm 2.47$ & $82.96 \pm 2.47$ & $82.97 \pm 2.49$\\
        Loan (accuracy)    & $78.03 \pm 4.91$ &  $78.36 \pm 4.32$ & $78.32 \pm 4.33$ & $78.54 \pm 5.08$\\
        Expense (MSE)  & $192.14 \pm 102.96$ & $243.49 \pm 107.51$ & - & -\\
        \bottomrule
    \end{tabular}
    \caption{Performance of all models. SaDe performs comparably to the baselines.}
    \label{tab:perf}
\end{table}

\subsubsection{Q2} 
Does SaDe's restriction to admissible models affect predictive performance? 
Table \ref{tab:perf} compares the predictive performance (accuracy / MSE on test data that do not violate constraints) of the learned models. 
For Loan Approval, SaDe performs comparably with the baselines. For Music Genre Identification, it performs slightly worse, while for Expense Prediction it performs better. The differences are not significant though. 


These results show that SaDe has the potential to return admissible models without a substantial cost to predictive performance.


\begin{table}[t]
    \centering
    \begin{tabular}{@{}lrrrr@{}}
        \toprule
        \textbf{Use-case} &  \textbf{SaDe} &  \textbf{SBR} &  \textbf{SL} &  \textbf{PP} \\
        \midrule 
        Music Genre     &  $3146 \pm 962$ &  $519 \pm 38$ & $513 \pm 42$ & $515 \pm 39$\\
        Loan Approval    & $134 \pm 66$ &  $85 \pm 3$ & $86 \pm 3$ & $88 \pm 3$\\
        Expense Prediction  & $3297 \pm 514$ & $123 \pm 53$ & - & -\\
        \bottomrule
    \end{tabular}
    \caption{SaDe requires more modelling time than the regularisation-based models.}
    \label{tab:runtimes}
\end{table}

\subsubsection{Q3} Is there a price to pay in terms of learning time?
Table \ref{tab:runtimes} shows the run-times of SaDe and the baselines.
For these use-cases, SaDe takes about 2, 6, or 30 times longer to learn a model, compared to the regularisation based approaches.
This is not unexpected: SaDe solves the more complex task of not only finding models but also proving their admissibility. 

For safety-critical applications, such an increase in learning time would often be considered acceptable, given the guarantees one gets in return. Where this is not the case, there is room for investigating variants of SaDe that are potentially faster. For instance, SaDe's stopping criterion was not optimized in this work; a more sophisticated criterion might make the approach considerably faster. Also, recent advances in developing SMT solvers capable of verifying neural networks \cite{katz2019marabou} suggest that improvements in SMT solver technology may also positively affect SaDe's computational efficiency.



\subsubsection{The Post Processing Baseline}

We should devote some discussion to the post-processing baseline PP.
For the classification uses cases, PP works well: the combination of model and post-processing step satisfies the domain constraints (\cref{tab:adi}) with a similar performance (\cref{tab:perf}) as the baselines. For these specific cases, SaDe does not have an advantage over PP.

However, it is important to realize that PP is not a generally applicable approach.  It only works when there exists a trivial way to fix an individual prediction. 
For instance, when domain constraints enforce relationships between multiple targets, this kind of approach is not feasible.
This is showcased in the expense prediction use case.
The constraint ``{\it sum of all expenses must be smaller than the household income}'' does not translate to constraints on individual expenses, and there is more than one way 
in which individual predictions can be fixed in order to satisfy the domain constraint.  Even if a fixed procedure were introduced (e.g., reduce all of them proportionally), other constraints may interfere with this procedure, rendering it invalid; ``{\it going-out expense must not be more than 5\% of the household income}'' is such a constraint.




\subsection{Limitations of SaDe}
Our current implementation of SaDe still has a number of limitations.
Learning models with high degree of non-linearity (e.g. Neural Nets) has not been feasible up till now. The solver technology we are using was either too slow or its reasoning engine was simply too weak to be able to solve such problems. Future improvements in solver technology may make it possible to learn more complicated models using SaDe.

Additionally, our approach is not directly applicable to discrete models (e.g. Decision Trees) because SaDe relies on a differentiable loss function. A possible solution to this could be based on the ideas in Norouzi et al. \cite{NIPS2015_1579779b}: they learn a decision tree as a parametric model by approximating the global non-differentiable loss with a differentiable one. Such an approach could be explored in conjunction with SaDe.

\section{Conclusion}
We proposed a new learning framework based on maximum satisfiability and a novel learning algorithm $SaDe$ that can learn parametric models that provably satisfy user-provided domain constraints.
The framework is general enough to handle a wide range of learning problems (classification, regression, \ldots) and constraints.  
To our knowledge, our approach is the first to guarantee  admissibility of learned models for such a wide class of symbolically expressible constraints. While the approach is in principle generic and does not depend on the format of the model (as long as it has continuous parameters), there may be practical hurdles for complex model formats. We have empirically shown that the approach is feasible at least for linear models, that it guarantees admissibility where other approaches do not, and that this is often possible without a cost in predictive performance and acceptable cost in terms of training time. This makes the approach very relevant in application contexts that are safety-critical, governed by law or  company policies, etc. Our approach is just a first step in a direction in which there is much opportunity for further work. 

\appendix

\bibliographystyle{splncs04}
\bibliography{main}

\end{document}